# Managing Inconsistent Intelligence

Johan Schubert
Department of Data and Information Fusion
Division of Command and Control Warfare Technology
Defence Research Establishment
SE–172 90 Stockholm, Sweden
schubert@sto.foa.se
http://www.foa.se/fusion/

**Abstract -** *In this paper we demonstrate that it is possible to manage intelligence in constant time as a pre-process to information fusion through a series of processes dealing with issues such as clustering reports, ranking reports with respect to importance, extraction of prototypes from clusters and immediate classification of newly arriving intelligence reports. These methods are used when intelligence reports arrive which concerns different events which should be handled independently, when it is not known a priori to which event each intelligence report is related. We use clustering that runs as a back-end process to partition the intelligence into subsets representing the events, and in parallel, a fast classification that runs as a front-end process in order to put the newly arriving intelligence into its correct information fusion process.*

## 1 Introduction

In this paper we develop a method based on a series of processes for managing inconsistent intelligence information as a pre-process to information fusion. These processes taken together will classify any new incoming intelligence as belonging to a certain subproblem. In a series of back-end processes the intelligence is organized according to subproblem by a new very fast neural clustering method [1] capable of handling intelligence in large scale problems. This method combines the theory of Potts spin [2] with evidential reasoning [3–10].

In an earlier article [11] we developed a method using a neural network structure similar to the Hopfield and Tank model [12] for partitioning intelligence into clusters for relatively large scale problems. This clustering approach represented a great improvement in computational complexity compared to a previous method based on iterative optimization [13–17], although its clustering performance was not equally good. In order to improve clustering performance a hybrid of the two methods was also developed [18]. The neural clustering method was further extended [19] for simultaneous clustering and determination of number of clusters during iteration in the neural structure.

For very large problems we need a method with still lower computational complexity than achieved so far. This method is described in Section 4.3 and further explained in detail in a forthcoming article [1].

Based on the result of the clustering process a limited set of prototypes may be selected for future classification [20], in order to speed up computation in the time critical classification of incoming intelligence.

In Section 2 we describe how the conflict in Dempster's rule is used as an indicator of inconsistent information among two or more intelligence reports. In Section 3 we give an overview of the combined process. The individual processes are more completely described in Section 4.

## 2 The internal conflict of intelligence

When we fuse several intelligence reports we might notice that some of the information is not entirely consistent. Such inconsistencies can have several sources, but regardless of the reason for the inconsistency it is always an alarm bell.

We use evidential reasoning to handle the uncertainty of intelligence reports. In evidential reasoning mass is assigned by a basic probability assignment $m$ to a subset $A$ of an exhaustive set of mutually exclusive possibilities, a frame of discernment $\Theta$.

The internal conflict is a measure of the inconsistency among intelligence reports reporting different things that are deemed to be conflicting. Between a pair of intelligence reports the conflict is the sum of all products of support of logically inconsistent statements, e.g., if our first intelligence report says $A$, $B$ or $C$, while our second report concerning the same issue says $D$ or $E$, where $A$ and $D$, as well as $B$ and $E$, are deemed to be inconsistent statements, i.e., $A \cap D = B \cap E = \varnothing$, then the conflict between these two reports is $m(A)m(D) + m(B)m(E)$. This number is between zero and one. The higher this value is, the more conflict there is between the two intelligence reports. When the conflict is one the two reports are completely inconsistent, while a conflict of zero is no indication that the two reports belong together in some way, it is merely a lack of inconsistent information. Thus, the conflict is always a form of negative indication and the lack of a negative indication is no positive information, it is just not negative. This realization will turn out to be important when developing methods to handle the inconsistency among several intelligence reports.

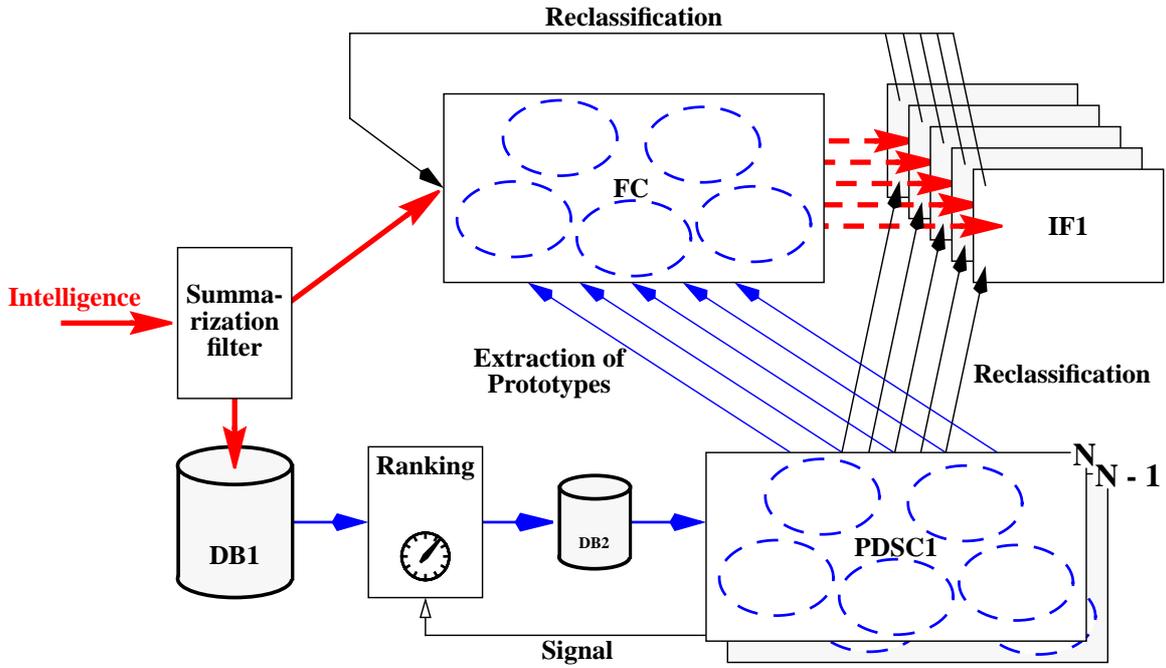

Figure 1: The intelligence management process.

## 3 Process overview

In order to achieve fast classification of incoming intelligence the intelligence will be handled by two parallel lines of reasoning, Figure 1.

The first line of reasoning is based on obtaining a fast classification that immediately puts the intelligence into the correct information fusion process. Before that, however, a summarization filter is passed where focal elements with very little support are eliminated. This puts a restriction on the maximal size of one intelligence report, assuring that the classification may execute in constant time for any intelligence report.

The second line of reasoning is performed in parallel with the first. Here, new intelligence is clustered with old in order to find a best partitioning of all intelligence into subsets representing different events that can be handled independently. Determining the number of subsets can be a difficult task. As the process runs continuously we will do one clustering with the same number of subsets as last time (PDSC1), and a second clustering (PDSC2) in parallel with one subset less. After each clustering process is finished we compare the conflict in each subset with a predetermined threshold.

Such a threshold can be found by careful experimentation when it is known that only one event is taking place. The received conflict in such an experiment is considered acceptable and can serve as the needed threshold. This approach was used by a European industrial defence company [21].

We will use the result of the first clustering (PDSC1) unless the largest conflict among all subsets of the second clustering (PDSC2) is less than the threshold. If it is, we will instead use the result of the second clustering. The same number of subsets used in PDSC2 this time will then be used in PDSC1 the next. If on the other hand the largest conflict among all subsets of PDSC1 is larger than the threshold, we will still use that result now, but for next time we increase the number of subsets for PDSC1 by one. While waiting for the next run of the clustering processes all intelligence is stored in a database (DB1).

In order for the two clustering processes to run in constant time a limited number of intelligence reports are allowed to take place in the process. This limit is typically domain dependent. By using a ranking method to evaluate reports based on information content we may find the most informative set of intelligence reports. These are put into DB2 where they are found by the clustering process.

Finally, a limited number of representative prototypes are extracted from the result of one clustering process in order to speed up the fast classification (FC) further.

After classification of incoming intelligence, the intelligence can be fused in the information fusion process (IF1, ...) where it belongs.

## 4 The Processes

### 4.1 The filtering pre-process

The filtering pre-process is a summarization filter that limits the number of focal elements in a piece of evidence, i.e., intelligence report. The idea is simple: any focal element other than the frame itself, $\Theta$, with a basic probability assignment below a certain threshold $p_0$ is eliminated, and its mass is reassigned to the entire frame.

For example, if $m_1$ is a piece of evidence with three focal elements $A$, $B$, and $\Theta$ coming in to the filter, where $m_1(B) < p_0 \leq m_1(A)$, then $m_1$ is exchanged for $m_1^*$, where $m_1^*(A) = m_1(A)$, $m_1^*(\Theta) = m_1(\Theta) + m_1(B)$.

As output from the filter, $m_1^*$ is directly sent on to the classifier for immediate classification, as well as to the database DB1 from where it might later be selected to influence the clustering process and the classifier itself. The incoming piece of evidence $m_1$ is eliminated.

By using this simple filter we limit the number of focal elements per piece of evidence and assure ourselves that an orthogonal combination of two pieces of evidence can always by performed in constant time.

For instance, if $p_0 = 0.01$ there are a maximum of 100

focal elements in every piece of evidence that has passed through the filter. Thus the combination of any two pieces of evidence has a maximum number of operations and can be performed in constant time.

## 4.2 The ranking process

In order for the clustering process in the next step to be able to run in constant time we must limit the total number of pieces of evidence taking part in the upcoming clustering process.

The number taking part in the clustering must be decided by the actual application and its time constraints. At this time before any clustering has taken place, it is not possible to rank pieces of evidence based on their real usefulness, or some other similar criteria. Instead, we will rank the evidence based on their potential usefulness as measured by their average total uncertainty [22, 23] within each body of evidence. This is measured as the sum of Shannon entropy and Hartley information for each piece of evidence $m$, which measures both scattering and nonspecificity:

$$H(m) = -\sum_{A \in \Theta} m(A)\log\{m(A)\} + \sum_{A \in \Theta} m(A)\log(|A|). \quad (1)$$

We choose the decided number of pieces of evidence ranked according to minimal average total uncertainty $H(m)$. Obviously, using too small a number of chosen pieces of evidence creates the risk of having a nonrepresentative clustering result and thus in the end an erroneous, albeit fast classification.

If $A$ is made time dependent such that $|A| \to \infty$ as $t \to \infty$ the measure $H(m)$ can take into account the decreasing value of old intelligence over time.

## 4.3 The clustering process

We consider the case when evidence come from multiple events which should be handled independently, and it is not known to which event a piece of evidence is related. We use the clustering process to separate the evidence into subsets for each event, so that each subset may be handled separately.

We combine Dempster-Shafer theory with the antiferromagnetic Potts model [2] into a powerful solver for very large Dempster-Shafer clustering problems [1]. We believe this method can serve as a general solution for preprocessing of intelligence data in information fusion.

The Hopfield model [24], based on the physics of Ising spins [25], was the first model to bridging the gap between spin systems and computer science that gained a wider interest. The Potts model [2] is a generalization of the Ising model where each spin may have an arbitrary (but finite) degree of freedom instead of just two. It has proven useful in many complex optimization problems [26].

If the Potts spin at a site $i$ is denoted $\sigma_i = 1, 2, ..., q$, where $q$ is a positive integer, the energy function that defines the model is written in terms of spin-spin interactions,

$$E = \frac{1}{2}\sum_{i,j=1}^{N} J_{ij}\delta_{\sigma_i\sigma_j}. \quad (2)$$

Another useful notation is to treat each Potts spin as a discrete vector in a hypercube: $S_{ia} = 0, 1$ with the constraint $\sum_{a=1}^{q} S_{ia} = 1\ \forall i$, where $a$ is a vector index. Then the energy function becomes

$$E = \frac{1}{2}\sum_{i,j=1}^{N}\sum_{a=1}^{q} J_{ij}S_{ia}S_{ja} \quad (3)$$

The spins merely encode which class a data (point) belongs to; $S_{ia} = 1$ means that the site $i$ belongs to class $a$.

This model can serve as a data clustering algorithm with a spin on each data point (site), if $J_{ij}$ is used as a penalty factor of site $i$ and $j$ being in the same class; sites in different classes get no penalty.

The problem consists of minimizing an energy function by flipping the spins into different states. This spin flipping process takes place via simulated annealing. At a high temperature, the spins flip more or less at random, and are only marginally biased by their interactions ($J_{ij}$). As the temperature is lowered parts of the system become constrained in one way or the other, they freeze. Finally, when the complete system is frozen, the spins are completely biased by the interactions ($J_{ij}$) so that, hopefully, the minimum of the energy function is reached. For computational reasons we will use a mean field model, where spins are deterministic [26].

We want to partition the set of all pieces of evidence $\chi$ into subsets where each subset refers to a particular event. These subsets are denoted by $\chi_i$. The conflict when all pieces of evidence in $\chi_i$ are combined by Dempster's rule is denoted by $c_i$. We can use the conflict in Dempster's rule when all pieces of evidence within a subset are combined as an indication of whether these pieces of evidence belong together. The higher this conflict is, the less credible that they belong together.

In [13] a criterion function of overall conflict called the metaconflict function for reasoning with multiple events was established. The metaconflict is derived as the plausibility that the partitioning is correct when the conflict in each subset is viewed as a piece of metalevel evidence against the partitioning of the set of evidence, $\chi$, into the subsets, $\chi_i$.

DEFINITION. *Let the* metaconflict function,

$$Mcf(q, S_1, S_2, ..., S_n) \triangleq 1 - \prod_{i=1}^{q}(1 - c_i), \quad (4)$$

*be the conflict against a partitioning of n pieces of evidence of the set $\chi$ into q disjoint subsets $\chi_i$. Here, $c_i$ is the conflict in subset i.*

We will use the minimizing of the metaconflict function as the method of partitioning the evidence into subsets corresponding to the events. After this, each subset refers to a different event and the reasoning can take place with each event treated separately.

The metaconflict function is easier to treat if it is rewritten as a sum instead of a product, by taking the logarithm. Let us rewrite the minimization as follows

$$\min Mcf$$
$$\Leftrightarrow$$
$$\max \log(1 - Mcf) = \max \log \prod_i (1 - c_i) \quad (5)$$
$$= \max \sum_i \log(1 - c_i) = \min \sum_i -\log(1 - c_i)$$

where $-\log(1 - c_i) \in [0, \infty]$ is a weight [5] of evidence, i.e., in this context a weight of conflict.

Since the minimum of $Mcf$ ($= 0$) is obtained when the final sum is minimal (= 0), the minimization of the final sum yields the same result as a minimization of Mcf would have done.

In Dempster-Shafer theory one defines a simple support function, where the evidence points precisely and unambiguously to a single nonempty subset $A$ of $\Theta$. If $S$ is a simple support function focused on $A$, then the basic probability numbers are denoted $m(A) = s$, and $m(\Theta) = 1 - s$. If two simple support functions, $S_1$ and $S_2$, focused on $A_1$ and $A_2$ respectively, are combined, the weight of conflict between them is [5]

$$\text{Con}(S_1, S_2) = \begin{cases} -\log(1 - s_1 s_2), & \text{if } A_1 \cap A_2 = \varnothing \\ 0, & \text{else} \end{cases}, \quad (6)$$

which may be written as

$$\text{Con}(S_1, S_2) = -\log(1 - s_1 s_2) \delta_{|A_1 \cap A_2|} \quad (7)$$

with $\delta_{|A_1 \cap A_2|}$ being defined so that it is unity for $A_1 \cap A_2 = \varnothing$ and zero otherwise.

Using the vector notation for the Potts Spin, the complete energy function we are considering is

$$E[S] = \frac{1}{2} \sum_{a=1}^{K} \sum_{i,j=1}^{N} J_{ij} S_{ia} S_{ja} - \frac{\gamma}{2} \sum_{a=1}^{K} \sum_{i=1}^{N} S_{ia}^2 + \frac{\alpha}{2} \sum_{a=1}^{K} \left( \sum_{i=1}^{N} S_{ia} \right)^2 \quad (8)$$

where the first term is the standard clustering cost.

The Potts mean field equations are [26]:

$$V_{ia} = \frac{e^{-H_{ia}[V]/T}}{\sum_{b=1}^{K} e^{-H_{ib}[V]/T}} \quad (9)$$

where

$$H_{ia}[V] = \frac{\partial E[V]}{\partial V_{ia}} = \sum_{j=1}^{N} J_{ij} V_{ja} - \gamma V_{ia} + \alpha \sum_{j=1}^{N} V_{ja} \quad (10)$$

with $V_{ia} = \langle S_{ia} \rangle$, and are used recursively until a stationary equilibrium state has been reached for each temperature. To apply it to Dempster-Shafer clustering we use interactions $J_{ij} = -\log(1 - s_i s_j) \delta_{|Ai \cap Aj|}$.

The algorithm for simulating these spins works roughly as follows. Use a precomputed highest critical temperature, $T_c$, as the starting temperature. Choose the mean field spins to be in their symmetric high temperature state; $V_{ia} = 1/K \ \forall i, a$. At each temperature, iterate eqs. (9), (10) until a fix point has been reached. The temperature is lowered by a constant factor until every spin has frozen, i.e., $V_{ia} = 0, 1$, Figure 2.

On a test problem we clustered $2^K - 1$ pieces of evidence into $K$ subsets. The evidence supports all subsets of the frame $\Theta = \{1, 2, 3, \ldots, K\}$. Thus, there always exists a global minimum to the metaconflict function equal to zero.

The reason we choose a problem where the minimum metaconflict is zero is that it makes a good test example for evaluating performance.

We notice an exponential computation time in the number of items of clusters. This is solely due to the exponential growth in the number of items of evidence via $N = 2^K - 1$. Although $K$ (= $|\Theta|$; the number of clusters) and $N$ (= $|2^{\Theta}| - 1$; the number of items of evidence) are not changed independently in these test examples, evidence is rather striking that the Potts Spin computation time scales

**INITIALIZE**
  $K$ (the problem size); $N = 2^K - 1$;
  $J_{ij} = -\log(1 - s_i s_j) \delta_{|Ai \cap Aj|} \quad \forall i, j$;
  $s = 0; t = 0; \varepsilon = 0.001; \tau = 0.9; \alpha$ (for $K \le 7$: $\alpha = 0$, $K = 8$: $\alpha = 10^{-6}$, $K = 9$: $\alpha = 0$, $K = 10$: $\alpha = 3 \cdot 10^{-7}$, $K = 11$: $\alpha = 3 \cdot 10^{-8}$); $\gamma = 0.5$;
  $T^0 = T_c$ (a critical temperature) $= \frac{1}{K} \cdot max(-\lambda_{min}, \lambda_{max})$, where $\lambda_{min}$ and $\lambda_{max}$ are the extreme eigenvalues of M, where $M_{ij} = J_{ij} + \alpha - \gamma \delta_{ij}$;
  $V_{ia}^0 = \frac{1}{K} + \varepsilon \cdot rand[0,1] \quad \forall i, a$;

**REPEAT**
  • REPEAT–2
    $\forall i$ Do:
      • $H_{ia}^s = \sum_{j=1}^{N} (J_{ij} + \alpha) V_{ja}^s - \gamma V_{ia}^s \quad \forall a$;
      • $F_i^s = \sum_{a=1}^{K} e^{-H_{ia}^s / T^t}$;
      • $V_{ia}^{s+1} = \frac{e^{-H_{ia}^s / T^t}}{F_i^s} + \varepsilon \cdot rand[0,1] \quad \forall a$;
      • $s = s + 1$;
    UNTIL–2
      $\frac{1}{N} \sum_{i,a} |V_{ia}^s - V_{ia}^{s-1}| \le 0.01$;
  • $T^{t+1} = \tau \cdot T^t$;
  • $t = t + 1$;
**UNTIL**
  $\frac{1}{N} \sum_{i,a} (V_{ia}^s)^2 \ge 0.99$;

**RETURN**
  $\{\chi_a | \forall S_i \in \chi_a. \forall b \ne a \ V_{ia}^s > V_{ib}^s\}$;

Figure 2: the clustering algorithm.

as $N^2 \log^2 N$.

The Potts Spin method is able to find a global optimum for problem sizes up to nine clusters. However, for the ten- and eleven-cluster problems the metaconflict increases rapidly.

A large part of the increase in metaconflict is due to the increase in problem size. Each cluster contributes to the total metaconflict, and as the number of cluster increases, the total metaconflict increases as well. In order to eliminate this effect we must calculate the average metaconflict per cluster.

In the eleven-cluster problem fluctuation of the results is large. While the method is still able to find some near optimal partitioning the average partition yields a higher metaconflict per cluster. This is a clear indication that the Potts method has reached its limit to produce perfect solutions, but still produces near optimal solutions.

The best measure of clustering performance is the metaconflict per evidence. Simply divide the average metaconflict per cluster already found with the average number of pieces of evidence in each cluster. This way we also take into account the exponential growth in the number of items of evidence as the number of clusters grow. The remarkable result is that the Potts model does not give any significant rise of the mean metaconflict per evidence. This is

  

true for almost three orders of magnitude.

For the eleven-cluster problem the Potts method achieves mean and median metaconflicts per evidence of just 0.8‰ and 2.4‰, respectively.

The results show that on average the metaconflict per evidence obtained corresponds to much less than one conflicting pair of pieces of evidence per cluster. This must be considered to be a very good result.

## 4.4 The prototype extraction process

Although every piece of evidence was placed in the best subset for that piece of evidence by the clustering process, some pieces of evidence might belong to one of several different subsets. Such an item of evidence is not so useful and should not be used as a prototype.

We must find a measure of the credibility that it belongs to the subset in question. A piece of evidence that cannot possibly belong to a subset has a credibility of zero for that subset, while a piece of evidence which cannot possibly belong to any other subset and is without any support whatsoever against this subset has a credibility of one. That is, the degree to which some piece of evidence can belong to a certain subset and no other, corresponds to the importance it wields in that subset.

The *credibility* $\alpha_j$ of $e_q$ when $e_q$ is used in $\chi_j$ is calculated as

$$\alpha_j = \frac{[\text{Pls}(e_q \in \chi_j)]^2}{\sum_k \text{Pls}(e_q \in \chi_k)}. \qquad (11)$$

Each piece of evidence is a potential prototype for its most credible subset.

We see that maximizing the credibility $\alpha_j$ for $e_q$ is equal to minimizing $m(e_q \notin \chi_j)$ for all $j$.

A simple decision rule is then:
For every piece of evidence $e_q$ and all $j$ find the minimum $m(e_q \notin \chi_j)$. Now, we have $e_q$ as a potential prototype for $\chi_j$.

The value of $m(e_q \notin \chi_j)$ is found by observing changes in cluster conflicts when a piece of evidence is moved from one subset to another, Figure 3.

If a piece of evidence $e_q$ in $\chi_j$ is taken out from the subset, the conflict $c_j$ in $\chi_j$ decreases to $c_j^*$. This decrease $c_j - c_j^*$ is interpreted as evidence indicating that $e_q$ does not belong to $\chi_j$. After some calculations we find

$$m(e_q \notin \chi_j) = \frac{c_j - c_j^*}{1 - c_j^*}. \qquad (12)$$

However, we must use the credibility itself when determining which pieces of evidence among the potential prototypes for a certain subset will actually by chosen as one of $N$ prototypes for that subset. While the above approach chooses the best subset $\chi_j$ for each piece of evidence by minimizing $m(e_q \notin \chi_j)$, it is still possible that the evidence might be quite useless as a prototype since it could almost have been a potential prototype for some other subset. By ranking the potential prototypes within each subset according to credibility, we are able to find the most appropriate ones.

We choose the $N$ prototypes with the highest credibility for the subset. They are judged to be the best representatives of the characteristics of the subset. Our future classification will be based on a comparison between them and the new incoming piece of evidence.

A second decision rule can then be formulated as:
For every subset:
1. Of the potential prototypes allocated for a subset, choose the $N$ prototypes with highest credibility for that subsets as the actual prototypes.
2. Disregard the other potential prototypes for that subset.

By first applying the first decision rule for all pieces of evidence and then the second decision rule for every subset we are able to find $N$ different prototypes for each subset provided, of course, that there are at least $N$ potential prototypes for each subset.

Finally, we combine all prototypes within each subset into one new basic probability assignment. This way, each subset will now contain only one piece of evidence. While doing that, we also make a note of the conflict $c_j$ received in that combination. We will need it in the fast classification process.

## 4.5 The fast classification process

Now, given that we have all the prototypes for each subset, we can make a fast classification of future incoming pieces of evidence. We will use the derived items of metalevel evidence $m(e_q \notin \chi_i)$.

If the evidence for $e_q$ against every subset is very high we will not classify $e_q$ as belonging to any of the subsets $\chi_j$. We will use a rejection rule if the best subset for $e_q$ brings a conflict higher than the threshold (see Section 3).

Our rejection rule is:
Reject $e_q$ if the minimum for all $j$ of $m(e_q \notin \chi_j)$ is larger than the threshold, where

$$m(e_q \notin \chi_j) = \frac{c_j^* - c_j}{1 - c_j}. \qquad (13)$$

If $e_q$ is not rejected by this rule, then $e_q$ is classified as belonging to the subset $\chi_j$ for which $m(e_q \notin \chi_j)$ is minimal for all $j$.

All it takes to find $c_j^*$ is one combination of Dempster's rule for each cluster between the incoming piece of evidence and the already made combination of the prototypes of that cluster.

If a fixed maximum number of prototypes are used for each cluster then the classification can always be done in a constant time. That is, independent of the total number of pieces of evidence in the previous clustering process.

## 4.6 The information fusion process

Having been classified by the pre-process and put into the best subset the new intelligence can now be fused with old intelligence that are already in that subset. Since the intelligence in any subset is independent of all intelligence outside of the subset, each subset can now be handled independently in a separate information fusion process.

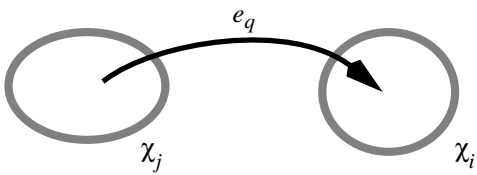

Figure 3: Moving a piece of evidence changes the conflicts in $\chi_i$ and $\chi_j$.

These fusion processes run in parallel.

As intelligence grows older we may find some unnecessarily large conflict building up over time in some of the information fusion processes. At such time, it may be important to take a second look on the partitioning and reclassify the already existing intelligence inside the different fusion processes. This is done at the time when that clustering process terminates by using the result as a reclassification of those intelligence reports inside the fusion processes that took part in the clustering. For those newly arrived intelligence reports that arrived to the fusion processes after the last clustering process was started another approach is needed. They may now be reclassified in the same way as any new arriving intelligence.

## 5    Conclusions

Through a series of moderate to small approximations we have shown that it is possible to manage inconsistent intelligence in constant time and have a very fast classification of all incoming intelligence in a pre-process before information fusion takes place.